\documentclass[10pt,twocolumn,letterpaper]{article}

\usepackage{iccv}
\usepackage{times}
\usepackage{epsfig}
\usepackage{graphicx}
\usepackage{amsmath}
\usepackage{amssymb}
\usepackage{algorithm}
\usepackage{algorithmicx,algpseudocode}

\usepackage[breaklinks=true,bookmarks=false]{hyperref}

\iccvfinalcopy 

\def\iccvPaperID{****} 

\ificcvfinal\fi
\begin{document}

\title{Large-Scale 3D Shape Reconstruction and Segmentation from ShapeNet Core55}

\author{Li Yi$^1$ \quad Lin Shao$^1$ \quad Manolis Savva$^2$ \quad Haibin Huang$^3$ \quad Yang Zhou$^3$ \quad Qirui Wang$^4$\\
Benjamin Graham$^5$ \quad Martin Engelcke$^6$ \quad Roman Klokov$^7$ \quad Victor Lempitsky$^7$ \quad Yuan Gan$^8$\\
Pengyu Wang$^8$ \quad Kun Liu$^8$ \quad Fenggen Yu$^8$ \quad Panpan Shui$^8$ \quad Bingyang Hu$^8$ \quad Yan Zhang$^8$\\
Yangyan Li$^9$ \quad Rui Bu$^9$ \quad Mingchao Sun$^9$ \quad Wei Wu$^9$ \quad Minki Jeong$^{10}$ \quad Jaehoon Choi$^{10}$\\
Changick Kim$^{10}$ \quad Angom Geetchandra$^{11}$ \quad Narasimha Murthy$^{11}$ \quad Bhargava Ramu$^{11}$\\
Bharadwaj Manda$^{11}$ \quad M Ramanathan$^{11}$ \quad Gautam Kumar$^{13}$ \quad Preetham P$^{13}$\\
Siddharth Srivastava$^{13}$ \quad Swati Bhugra$^{13}$ \quad Brejesh Lall$^{13}$ \quad Christian H{\"a}ne$^{14}$\\
Shubham Tulsiani$^{14}$ \quad Jitendra Malik$^{14}$ \quad Jared Lafer$^{15}$ \quad Ramsey Jones$^{15}$ \quad Siyuan Li$^{16}$\\
Jie Lu$^{16}$ \quad Shi Jin$^{16}$ \quad Jingyi Yu$^{16}$ \quad Qixing Huang$^{17}$ \quad Evangelos Kalogerakis$^3$\\
Silvio Savarese$^1$ \quad Pat Hanrahan$^1$ \quad Thomas Funkhouser$^2$\quad Hao Su$^{12}$ \quad Leonidas Guibas$^1$\\
\small $^1$Stanford University \enskip $^2$Princeton University \enskip $^3$University of Massachusetts--Amherst \enskip $^4$Tsinghua University \enskip $^5$Facebook AI Research\\
\small $^6$University of Oxford \enskip $^7$Skolkovo Institute of Science and Technology \enskip $^8$Nanjing University \enskip $^9$Shandong University\\
\small $^{10}$Korea Advanced Institute of Science and Technology \enskip $^{11}$Indian Institute of Technology Madras \enskip $^{12}$UC San Diego\\
\small $^{13}$Indian Institute of Technology Delhi \enskip $^{14}$UC Berkeley \enskip $^{15}$Imbellus \enskip $^{16}$ShanghaiTech University \enskip $^{17}$University of Texas, Austin
}

\maketitle
\thispagestyle{empty}

\begin{abstract}
   We introduce a large-scale 3D shape understanding benchmark using data and annotation from ShapeNet 3D object database. The benchmark consists of two tasks: part-level segmentation of 3D shapes and 3D reconstruction from single view images. Ten teams have participated in the challenge and the best performing teams have outperformed state-of-the-art approaches on both tasks. A few novel deep learning architectures have been proposed on various 3D representations on both tasks. We report the techniques used by each team and the corresponding performances. In addition, we summarize the major discoveries from the reported results and possible trends for the future work in the field.
\end{abstract}

\section{Introduction}
The goal of this challenge is to synergize the efforts of the broad community (Computer Graphics, Computer Vision, Machine Learning) related with 3D shape processing. Recently, the shape processing community is being transformed with the emergence of big 3D dataset and the increasing usage of learning based approaches. In particular, papers published in the past 3 years have shown that deep learning methods are quite effective for both understanding and synthesizing 3D content. Along with the shift of technical trend, the general interests are also expanding from solving traditional tasks, such as shape retrieval and multi-view reconstruction, to a broader range of tasks that aim to analyze 3d shapes or link 3D shapes with other modalities such as text and images. For example, in recent visual computing conferences, we witness papers on tasks such as object classification, part segmentation, novel-view synthesis, single-image based 3D reconstruction, shape completion, text to 3D scene, etc. 

Though more research directions in 3D data processing are flourishing, there lack well-designed and generally accepted benchmarks to fairly compare different approaches on the broad set of 3D tasks. As we have seen in image recognition community, challenges such as PASCAL VOC and ImageNet ILSVRC played a critical rule in synergizing the efforts from the entire community, by providing clear motivation and solid metrics to assess the progress of object recognition for researchers across industry and academia. In the field of 3D shape processing, traditional renowned benchmarks have mostly been focusing on 3D retrieval (SHREC); though a few benchmarks on other tasks are published along with technique papers, their test datasets are often small in scale and the evaluation metrics are often different from each other. 

We believe that it is timely for the community to introduce new large-scale 3D shape understanding benchmarks. In this first challenge organized by ShapeNet team, we pick two tasks among the few -- part-level 3D object segmentation and single-image based 3D object reconstruction. Our main criteria include the importance of the task and the readiness of the training data. Regarding task importance, 3D part segmentation is a key module in many applications, such as object manipulation, animation, geometric modeling, manufacturing; and single-image based 3D reconstruction evaluates machine's intelligence to infer 3D geometry from 2D images, an ability that is trivial for humans but extremely challenging for today's machines. Regarding data adequacy, ShapeNet database has already included adequate 3D object instances with necessary annotations to drive the development of learning-based methods. In the future, we hope to extend the benchmark to include more 3D tasks and augment the dataset with sensor data, beyond synthetic data from human modelers. 

In our challenge, training data with groundtruth annotations have been released publicly, while testing data are released without groundtruth annoations. Three weeks are given to all teams to work on one of the tasks and required to submit a description of their methods. In total 17 teams have registered the challenge and 10 teams have submitted their results. The best performing teams have outperformed baselines by our implementation of state-of-the-art methods on both tasks. 

As a summary of all approaches and results, we have the following major observations: 1) Approaches from all teams on both tasks are deep learning based, which shows the unparallel popularity of deep learning for 3D understanding from big data; 2) Various 3D representations, including volumetric and point cloud formats, have been tried. In particular, point cloud representation is quite popular and a few novel ideas to exploit point cloud format have been proposed; 3) The evaluation metric for 3D reconstruction is a topic worth further investigation. Under two standard evaluation metrics (Chamfer distance and IoU), we observe that two different approaches have won the first place. In particular, the coarse-to-fine supervised learning method wins by the IoU metric, while the GAN based method wins by the Chamfer distance metric. We hope that future researchers on relevant problems will draw inspirations and learn lessons from this challenge.

\section{Dataset}
In this challenge, we use 3D shapes from ShapeNet \cite{chang2015shapenet} to evaluate both segmentation and reconstruction algorithms. ShapeNet is an ongoing effort to centralize and organize 3D shapes online. 3D models are classified into different categories, which aligns with WordNet \cite{miller1995wordnet} synsets (lexical categories belonging to a taxonomy of concepts in the English language). ShapeNetCore contains a subset of ShapeNet models from 55 categories. These models are pre-aligned into a canonical frame in a category-wise manner.

In 3D shape part segmentation track, we use a subset of ShapeNetCore containing 16,880 models from 16 shape categories \cite{yi2016scalable}. Each category is annotated with 2 to 6 parts and there are 50 different parts annotated in total. 3D shapes are represented as point clouds uniformly sampled from 3D surfaces. Part annotations are represented as point labels, ranging from 1 to the number of parts for each shape. The task is defined as predicting a per point part label, given 3D shape point clouds and their category labels as input. To make training and evaluation of learning based methods possible, we establish training, validation and test splits of the dataset following the official splits for ShapeNetCore models. This results in 12136 training models, 1870 validation models and 2874 test models. There exists many nearly the same 3D models in ShapeNetCore. We remove 3D models in the test set with very similar geometry to a training or validation model. And we also remove approximate duplicated 3D models in the test set. These results in a test set with larger shape variation containing 1381 shapes. We evaluate all the segmentation approaches on both the ``original'' test set (2874 shapes) and the ``deduplicated'' test set with larger shape variation (1381 shapes).

In 3D shape reconstruction track, we use pre-aligned ShapeNetCore models which are split into 33673 training models, 4917 validation models and 10010 test models, summing up to 48,600 3D models in total. The task is defined as reconstructing a 3D shape, given a single image as input. We choose voxels as the output 3D representations. Each model is represented by $256^3$ voxels. Each voxel contains 1 or 0 as its value. 1 indicates occupancy and 0 indicates free space. Besides, we also provide the synthesized images as input. The synthesized images are rendered from random viewpoints sampled around textured 3D shapes. Category label for each shape is not provided for the reconstruction track. Similar to the segmentation task, we also generate a ``deduplicated'' test set with larger shape variation which contains 6785 test models. Again we evaluate all the reconstruction approaches on both the ``original'' test set (10010 shapes) and the ``deduplicated'' test set (6785 shapes). 

\section{Evaluation}
In 3D shape part segmentation track, participants are asked to submit their per point label prediction for each shape in the ``original'' test set. Average Intersection over Union (IoU) is used as the evaluation metric. Per part IoU is firstly computed for each part on each shape. Then per category IoU is computed for each category by averaging the per part IoU across all parts on all shapes with a certain category label. Finally an overall average IoU is computed through a weighted average of per category IoU. The weights are just the number of shapes in each category. The average IoU is computed both on the ``original'' test set and the ``deduplicated'' test set to compare different approaches. We compare different segmentation approaches in Section~\ref{seg_results}.

In 3D shape reconstruction track, participants are asked to submit their 3D reconstruction represented as $256^3$ voxels for each test image in the ``original'' test set. We then evaluate different approaches using two metrics, average Intersection over Union (IoU) and average Chamfer Distance (CD). Per shape IoU is firstly computed for each 3D reconstruction in the test set and then averaged to get the final IoU score. To compute the CD score, we firstly generate a point set representation for each shape by converting each occupied voxel into its central point. After this we adopt furthest point sampling to obtain a smaller point set containing $4096$ points for computation efficiency. Following the definition in \cite{fan2016point}, we could then compute the CD score for each 3D reconstruction and average across all test shapes to obtain the final score. Both the two evaluate metrics are computed for the ``original'' test set as well as the ``deduplicated'' test set. We show the comparison among different reconstruction approaches in Section~\ref{recon_results}.

\section{Participants}
There were seven participating groups in the segmentation track. Each group submitted their predicted shape labeling on the test shapes.
\begin{itemize}  
\item \textbf{SSCN}: Submanifold Sparse ConvNets, by B. Graham, M. Engelcke 
\item \textbf{PdNet}: Pd-networks, by R. Klokov, V. Lempitsky 
\item \textbf{DCPN}: Densely Connected Pointnet, by Y. Gan, P. Wang, K. Liu, F. Yu, P. Shui, B. Hu, Y. Zhang 
\item \textbf{PCNN}: PointCNN, by Y. Li, R. Bu, M. Sun, W. Wu 
\item \textbf{PtAdLoss}: PointAdLoss: Adversarial Loss for Shape Part Level Segmentation, by M. Jeong, J. Choi, C. Kim 
\item \textbf{KDTNet}: K-D Tree Network, by A. Geetchandra, N. Murthy, B. Ramu, B. Manda, M. Ramanathan 
\item \textbf{DeepPool}, by G. Kumar, P. P, S. Srivastava, S. Bhugra, B. Lall 
\end{itemize}

There were three participating groups in the reconstruction track. Each group submitted their shape reconstruction represented via $256^3$ voxels from test images.
\begin{itemize}
\item \textbf{HSP}: Hierarchical Surface Prediction, by C. H{\"a}ne, S.Tulsiani, J. Malik 
\item \textbf{DCAE}: Densely Connected 3D Auto-encoder by S. Li, J. Liu, S. Jin, J. Yu 
\item \textbf{$\alpha$-Gan}, by R.Jones, J.Lafer 
\end{itemize}

\section{Segmentation Methods}
In this section we compile the description of methods by each participating
group, from the participants’ perspective.

\subsection{Submanifold Sparse ConvNets, by B. Graham, M. Engelcke}
In a data preprocessing step, the point clouds are scaled according to their L2-norm and the points are voxelized into a sparse 3D grid. We augment the training data, for example by adding a small amount of noise to the location before voxelization. At size $50^3$, the voxels are 99\% sparse.

\begin{figure}[t]
\begin{center}
\includegraphics[width=0.75\columnwidth]{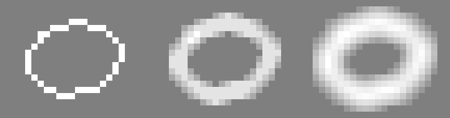}\\
(a) Regular sparse convolution.\\
\includegraphics[width=0.75\columnwidth]{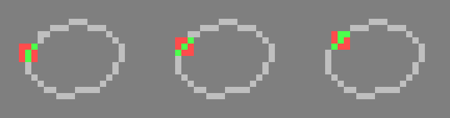}\\
(b) Valid sparse convolution.\\
\includegraphics[width=0.75\columnwidth]{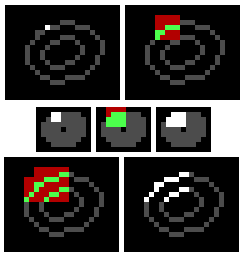}\\
(c) Block with a strided, a valid, and a de-convolution.\\
\end{center}
   \caption{Illustrations comparing (a) regular and (b) valid sparse convolutions, and (c) a computational block for increasing the receptive field while leaving the location of active sites unchanged. }
\label{fig:conv}
\end{figure}

We make use of sparse convolutional networks \cite{Graham2015} to solve the task.
The original formulation in \cite{Graham2015} was not well suited for semantic segmentation, as repeatedly applying size-3 convolutions blurs the set of active spatial locations (Fig. \ref{fig:conv} (a)), therefore diluting the sparsity and increasing the memory requirements.

We instead stack size-3 valid sparse convolutions (VSCs) \cite{Graham2017}. The filter outputs are only computed where the input locations are active (Fig. \ref{fig:conv} (b)), so the set of active sites remains unchanged. This is ideal for sparse segmentation.

To increase the receptive field, we add ResNet-style \cite{He2016} parallel paths with strided convolutions, VSCs, and deconvolutions (Fig. \ref{fig:conv} (c)). The illustration shows how one point in the input can influence many points in the output.

\subsection{Pd-networks, by R. Klokov, V. Lempitsky}
\subsubsection{Network Design}

\newcommand{\nd}[1]{n_{d}^{#1}}
\newcommand{\Wdl}[1]{^{#1}W_{d}^{l_i}}
\newcommand{\bdl}[1]{^{#1}\mathbf{b}_{d}^{l_i}}

\newcommand{\vt}{\tilde{\mathbf{v}}}
\newcommand{\Wtdl}[1]{^{#1}\tilde{W}_{d}^{l_i}}
\newcommand{\btdl}[1]{^{#1}\tilde{\mathbf{b}}_{d}^{l_i}}
\newcommand{\Sk}{S}
\newcommand{\tk}{\mathbf{t}}

Our model is a modification of Kd-Networks \cite{Klokov17} called principal direction networks (Pd-networks). Pd-networks work with modified PCA trees fit to point clouds (unlike common PCA trees, the trees in Pd-networks use small randomized subset to compute the principal direction at each node during recursive top-down tree construction). Like Kd-networks, Pd-networks can be used to classify point clouds. The recognition process performs a bottom-up pass over the pd-tree starting from some simple representations assigned to leaves (which correspond to individual points). Each computation during the bottom-up step computes the vectorial feature representation $\mathbf{v}_i$ for the parent node from the vectorial representations $\mathbf{v}_{c_k(i)}$ ($k=1,2$) of the children using the following operation:

{\small
\begin{align} \label{eq:pdnet_main}
    \mathbf{v}_i = \sum_{k=1}^2\sum_{d\in\{\mathtt{x},\mathtt{y},\mathtt{z}\}}\sum_{s=1}^2 {^s}\alpha_{d}^{k}(\Wdl{s}\phi(\mathbf{v}_{c_k(i)})+\bdl{s}),\\
    {^s}\alpha_{d}^{k} = \max(0;(-1)^{s}\nd{k}).\nonumber
\end{align}
}%
Here, the index $k$ iterates over the two children, the index $s$ looks at the positive and the negative components of the normal vectors, the index $d$ iterates over the three coordinates. Further, $i$ is the index for the parent node, $c_1(i) = 2i, c_2(i) = 2i + 1$ are the  node's children indices, $l_i$ is the parent node's depth level in the pd-tree, $\nd{k}$ is a normal associated with the split directed towards the $k$th child, $\phi$ is a non-linearity (a simple ReLU in our experiment), $\Wdl{s}, \bdl{s}$ are the trainable parameters (the matrix and the bias vector) of the Pd-network.

\begin{figure}
    \centering
    \includegraphics[width=\columnwidth]{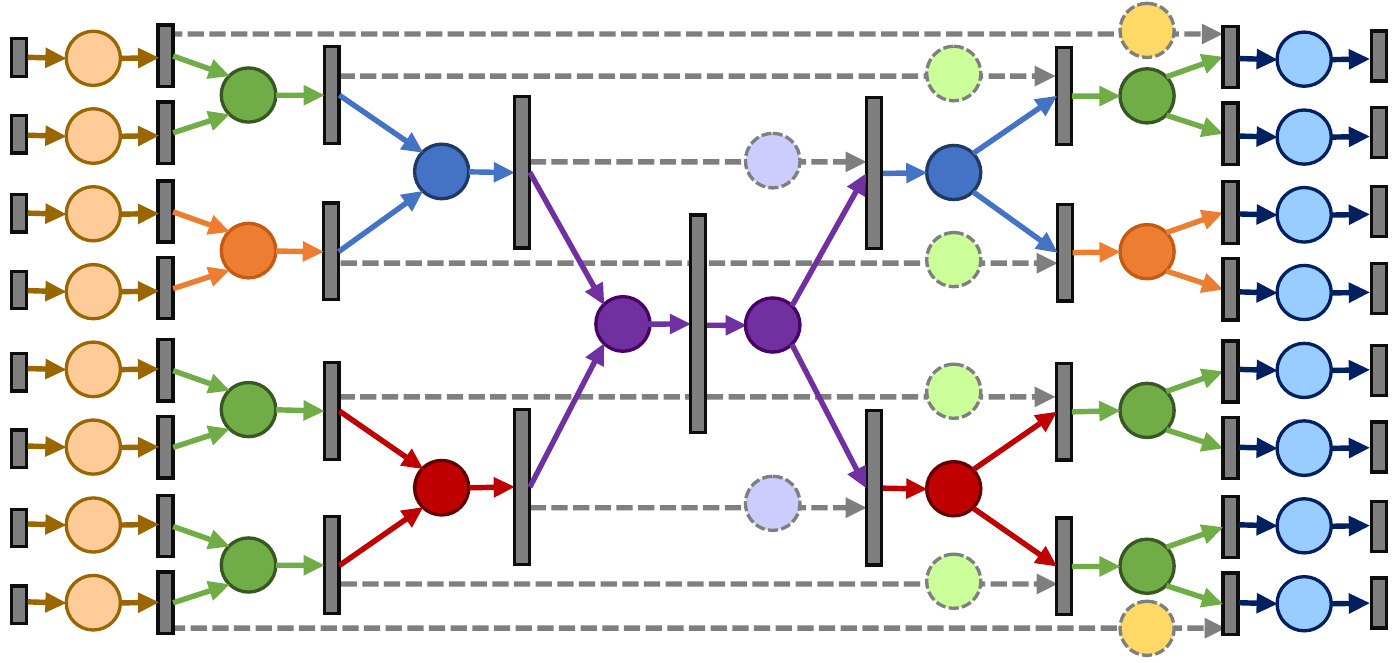}
    \caption{Pd-network for semantic parts segmentation. Arrows indicate computations that transform the representations (bars) of different nodes. Circles correspond to affine transformations followed by non-linearities. Similarly colored circles on top of each other share parameters. Dashed lines correspond to skip-connections (some ``yellow'' skip connections are not shown for clarity). The input representations are processed by an additional transformation (light-brown) and there are additional transformations applied to every leaf representation independently at the end of the architecture (light-blue).}
    \label{fig:pdsegm}
\end{figure}

The architecture for semantic segmentation is obtained from the classification architecture in the way that is analogous to \cite{Klokov17}. The bottom-up pass is inverted, with the top-down pass, where the new feature representation for children are computed from the representations of the parent nodes. Skip-connections between the bottom-up and the top-down passes are added leading to a standard U-net type architecture (Figure~\ref{fig:pdsegm}).

\subsubsection{Implementation details}
During training the model takes point clouds of size $4096$ (sampled from the original point clouds with an addition of a small noise) as inputs, compute pd-trees for the clouds and feed them to the network, which consists of: an affine transformation of size $128$ with parameters shared across all the points; an encoder-decoder part with node representation dimensionalities of sizes $128-128-128-128-256-256-256-256-512-512-512-512$ with one additional fully connected layer in the bottleneck of size $1024$; and two additional transformations at the end of the network of sizes $256, 256$ (the latter are applied at each leaf/point independently). During test time predictions are computed for the sampled clouds, then each point in the original cloud is passed through each of the constructed pd-trees to obtain a posterior over parts (the posterior distributions are averaged over multiple pd-trees). Small translations, anisotropic scaling along axes and computations of principal directions over randomized subsets ensure the diversity of pd-trees. In total, 16 pd-trees are used to perform prediction for each point cloud at test stage. The network was trained with adam optimizer with mini-batches of size 64 for 8 days. A forward pass performed on one pd-tree for a single shape takes approximately 0.2 seconds. A big advantage of Pd-networks for the segmentation task is their low memory footprint. Thus, for our particular architecture, the footprint of one example during learning is less than 140 Mb. 

\subsection{Densely Connected Pointnet, by Y. Gan, P. Wang, K. Liu, F. Yu, P. Shui, B. Hu, Y. Zhang}
\subsubsection{Approach Overview}
\begin{figure}[t]
\begin{center}
   \includegraphics[width=1.0\linewidth]{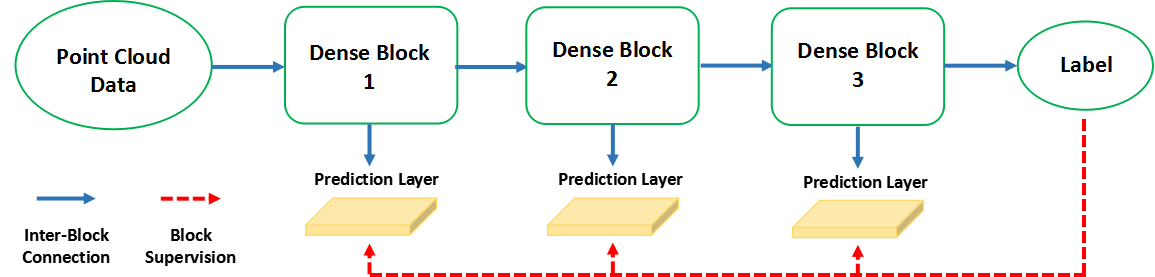}
\end{center}
   \caption{Three dense blocks are connected sequentially and each block has a supervision. Prediction layer is only a 1x1 convolution layer with {\em k} outputs.}
\label{fig:f1}
\end{figure}

\begin{figure}[t]
\begin{center}
   \includegraphics[width=1.0\linewidth]{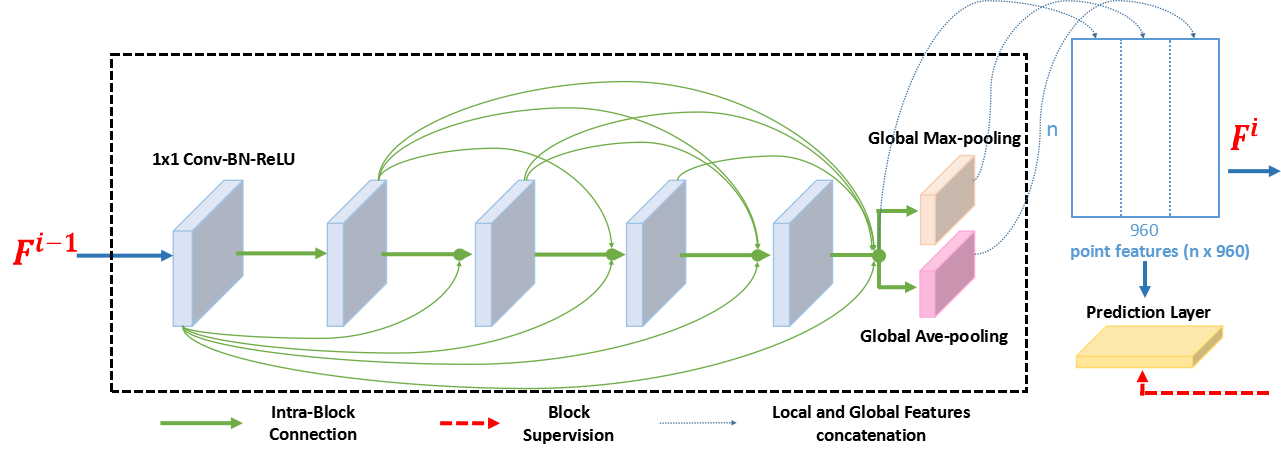}
\end{center}
   \caption{Five 1x1 convolution layers with 64 kernels have been used. All layers in the block are connected directly. Max-pooling and ave-pooling are used for extracting global features. The block's output Fi contains 960 dimensions point features which can be used for classification or as the input of next block's. }
\label{fig:f2}
\end{figure}

We desgin a novel network for this ShapeNet part segmentation challenge. Our network is illustrated in Fig~\ref{fig:f1}. This structure is inpired by Pointnet~\cite{Qi2017pointnet}, a simple but effective structure which combines local and global feaures to perform point label classification. We redesign a Pointnet~\cite{Qi2017pointnet} structure as a dense connection block like DenseNet~\cite{huang2017densely} and reduce the number of kernels. In every block, we connect all layers in the block directly and these local feaures are combined by concatenating. Meanwhile, to extract global features we use both max-pooling and ave-pooling feaures and concatenate them together. Then we concatenate all these local feaures and global features together as the per point features which can be used for label classification like Pointnet\cite{Qi2017pointnet} does. The structure details in a block are illustrated in Fig~\ref{fig:f2}. We use three blocks and each block has a supervision which can alleviates the difficulty in training deeper network and guarantees the discrimination of feature representations. After training, we use the dense conditional random field (CRF)~\cite{Philipp2013crf}to refine the results.

\subsubsection{Implementation Details}
We use the third prediction layer's output as the network result. In each block, five 1x1 convolution layers with 64 kernels have been used. Batchnorm and ReLU are used for all convolution layers and no DropOut. Prediction layer is only a 1x1 convolution layer with {\em k} outputs({\em k} means the number of parts of a category). Cross-entropy loss is minimized during training. We use adam optimizer with initial learning rate 0.001, momentum 0.90 and train the network on every category separately. It takes us around 30 minutes to train on a small category (e.g. bag, cap) and 2-3 hours to train on a large category (e.g. table, chair). We only use the original training data (around 3000 points with {\em (x,y,z)}    coordinate features per shape) and no data augmentation. In the CRF refining stage, we estimate the normal features from the given point cloud and use them as the input to minimize a energy function like~\cite{Wang2017ocnn} did. 

\subsection{PointCNN, by Y. Li, R. Bu, M. Sun, W. Wu}
\begin{figure}[t]
	\centering
	\includegraphics[width=1.0\columnwidth]{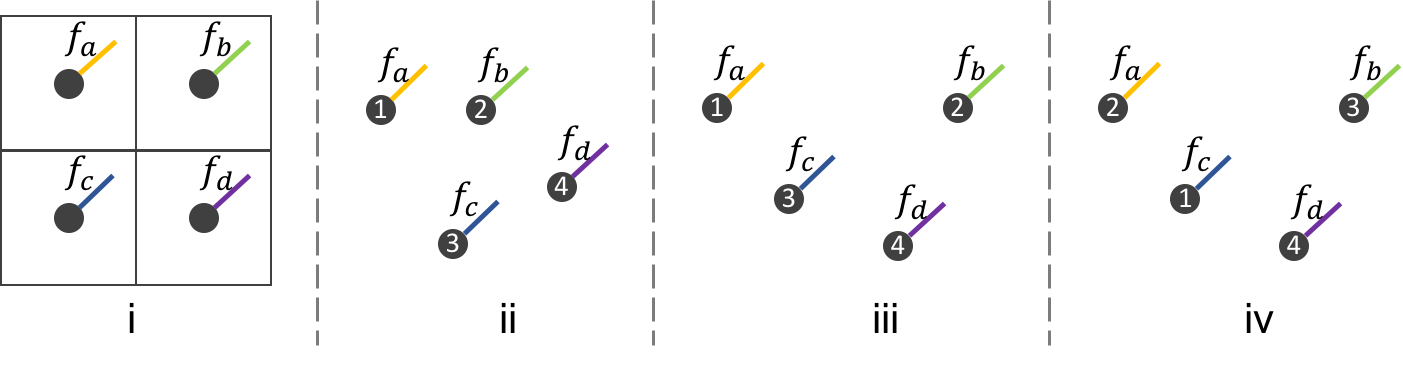}
	\caption{An illustration of input features $F=(f_a, f_b, f_c, f_d)^T$ from regular domain (i) and irregular domains (ii, iii, and iv). In all these cases, the set of the input features are the same. Suppose we have convolution kernels $T$. In (i), $F$ can be directly convolved with $T$, as they live in the same regular domain (can be considered as having same shape and order). In (ii) and (iii), the output features should be different, as the points (the indexed dots in the figures) are in different shapes, though the input features are the same. In (iii) and (iv), the output features should be same, as only the orders of the points differ, and the output features should be invariant to them. Clearly, directly convolving $F$ and $T$ in (ii, iii, iv) as that in (i) will not yield output features as desired.}
	\vspace{-0.4cm}
	\label{fig:motivation}
\end{figure}

Point cloud data, unlike images, is represented in irregular domains --- the points are arranged in different shapes and orders, each associated with some input features. To learn a feature representation for such data, clearly, the output features should take the point shapes into consideration (\emph{goal 1}), while being invariant to the point orders (\emph{goal 2}). However, a direct convolving of kernels against the features associated with the points, as that in typical CNNs, will result in deserting the shape information while being variant to the orders (see Figure~\ref{fig:motivation}). To address these problems, we propose to firstly transform the input features with respect to their point shapes while aiming for point order invariance, and then apply convolution on the transformed features. This is a general method for learning representations from point cloud data, and we call it \emph{PointCNN}. 

\paragraph{Problem Statement.} The input is a set of points each associated with a feature, i.e., $\mathcal{I}=\{(p_i, f_i)\}, i=1, 2, ..., K, p_i \in \mathbb{R}^3,$ and $f_i \in \mathbb{R}^{C_{in}}$. Without losing generality, we convert $\mathcal{I}$ into two matrices $(P, F)$ by given an arbitrary order to the points, where $P$ is a $K \times 3$ matrix storing the point coordinates, and $F$ is a $K \times C_{in}$ matrix storing the input features. The challenge is to design an operation $f$ for outputting feature $f_o \in \mathbb{R}^{C_{out}}$ that aims at the two goals mentioned before, i.e., $f: (P, F) \rightarrow f_o$.

\paragraph{PointConv.} Convolution operator $conv: F \rightarrow f_o$ has shown to be quite successful in learning representations, thus we opt to design $f$ based on $conv$. However, for $conv$ to be effective in our setting, the input features to $conv$ should take point shapes into consideration (goal 1), while being invariant to point orders (goal 2). Let us call the input features that aims at these goals $F_g$. Then the missing part in the design of $f$ is an operator $g: (P, F) \rightarrow F_g$, since we already have $conv: F_g \rightarrow f_o$.

For goal 1, we can think $F_g$ as a weighted sum of $F$, with the weights given by $P$, and different point shapes (different $P$s) give different weights. We realize this by learning a function $s: P \rightarrow W$ that maps the points $P$ into a $K \times K$ weighting matrix $W$, and then use $W$ to turn $F$ into $F_g$, i.e., $F_g = s(P) \times F$. $s$ is implemented with multi-layer perceptron. Interestingly, this $K \times K$ weighting matrix $W$, when approximating permutation matrix, can permute the features, thus resolve the point order issue and meet goal 2. In other word, $W$ can serve both goal 1 and 2, and since it is learned, it is up to the network and training for adaptively deciding the preference between goal 1 and 2. 

In a summary, PointConv $f$, the key build block of PointCNN, is defined to be a composition of two operators: $F_g = s(P) \times F$ for ``weighting'' and ``aligning'' input features,  followed by a typical convolution: $conv: F_g \rightarrow f_o$.

\paragraph{PointCNN.} Similar to convolution operator in CNN, PointConv can be stacked into PointCNN for learning hierarchical feature representations. PointConv is designed to process a set of points, maybe from a local neighborhood around a query point (KNN or radius search).   The query points for current layer are usually a downsampling of the input points, and will be the input points for the next layer, in sparse prediction tasks, such as classification.

\paragraph{PointCNN for Segmentation.} Since PointCNN is a natural generalization of CNN into processing point cloud data, the CNN architectures for various image tasks can be naturally generalized into processing point cloud data. We use ``Conv-DeConv'' style network for point cloud segmentation. The ``DeConv'' layers in PointCNN actually use exactly the same PointConv operator as that in ``Conv'' layers. The only difference is that the number of the query points are larger than the number of points being queried in the ``DeConv'' layers, for spatial upsampling.

\subsection{PointAdLoss: Adversarial Loss for Shape Part Level Segmentation, by M. Jeong, J. Choi, C. Kim}
\subsubsection{Network Design}
\begin{figure}[t]
    \begin{center}
    \includegraphics[width=0.88\linewidth]{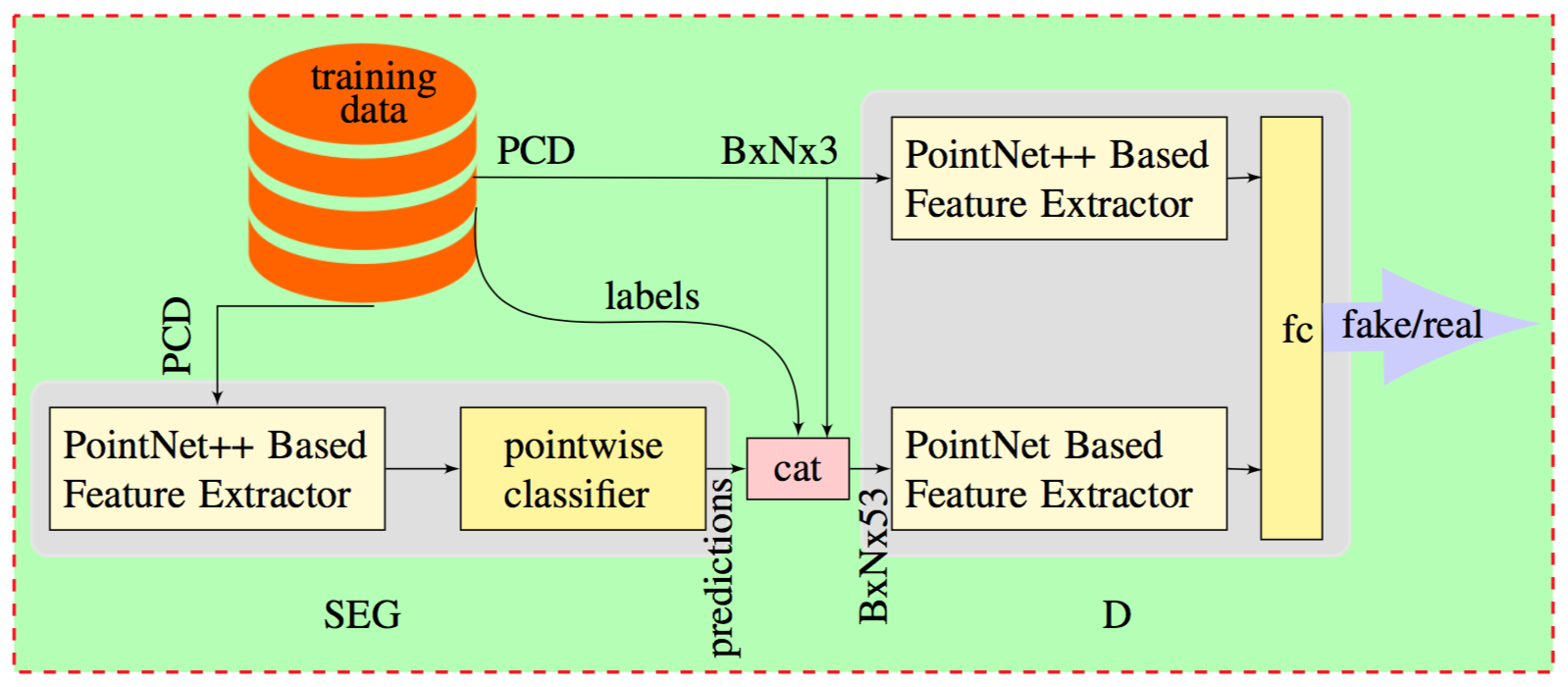}
    \caption{Detailed architecture of the networks.}
    \label{fig:net_architecture}
    \end{center}
\end{figure}

Based on notable works~\cite{Qi2017pointnet, qi2017pointnet++}, we design two deep networks for a part level segmentation task. One is a segmentation network, while the other is a discriminator network. The segmentation network takes point cloud data and computes the pointwise class probability. The discriminator network receives the data and the probability simultaneously and decides whether the probability is the ground truth label vector or the prediction vector. To get a pointwise feature from the data for the discriminator network, we concatenate a pointwise three-dimensional position vector and a $N$-dimensional class probability vector, where $N$ is the number of total part classes, $50$ in this challenge. Figure \ref{fig:net_architecture} describes the architecture of our networks.

\paragraph{Training Loss.} The networks are trained with two different losses: $\mathcal{L}_{SEG}$ for the segmentation network, and $\mathcal{L}_D$ for the discriminator network. We define the loss functions as follows.
\begin{equation}
    \mathcal{L}_{SEG} = CE(SEG(X), Y_{gt}) + \lambda D([X, SEG(X)], 1)
\end{equation}
\begin{equation}
    \mathcal{L}_D = D([X, SEG(X)], 0) + D([X, Y_{gt}], 1)
\end{equation}
where $X$ is the point cloud data, $Y_{gt}$ is the ground truth label, $CE$ indicates the cross entropy, $SEG(\cdot)$ refers to the class probability, $D(\cdot)$ implies the output of the discriminator network, $\lambda$ denotes a loss weight and $[\,,]$ means concatenation.

\subsubsection{Implementation Details}
We use TensorFlow for Python on a Linux environment to implement the networks. Since the discriminator network learns faster than the segmentation network, we add Gaussian noise to the concatenated input of the discriminator at the early stage of the training. Moreover, in order to slow down the training speed, we employ a training strategy to update the segmentation network twice and the discriminator network once for each training step. Our networks are trained for $10$ epochs. Both the learning rate of the segmentation network and the learning rate of the discriminator network are set to $0.001$.

The segmentation network consumes fixed $2,048$ points, even though the point cloud model could have a different number of points. We overcome this problem by dividing it into two cases. If the number of points in the model is smaller than $2,048$, we resample the points to make $2,048$ points. Otherwise, we divide the points into two subsets: i) randomly sampled $2048$ points, and ii) the rest of the points (note: a model does not exceed $4,096$ points). In order to make the second subset trainable, we recreate another model by adding randomly sampled points from the first subset to make it $2,048$ points. We run forward pass for both subsets to get the final segmentation results.

\subsection{K-D Tree Network, by A. Geetchandra, N. Murthy, B. Ramu, B. Manda, M. Ramanathan}
\subsubsection{Method Description}

Segmentation is done on an order-invariant representation of the input point cloud. This representation is obtained by constructing a k-d tree for the point clouds. The output is fed into a fully-convolutional network with skip connections. A separate network is trained for each category.

\paragraph{K-D Tree.} For each model, we construct a k-d tree in which the split direction is taken to be the one having largest span or range among the points. The points are split such that the two resulting children have the same number of points. In case of odd number of points, the median point (along the split direction) is duplicated and included in both the children. This leads to an increase in the number of points. The number of leaves of the resulting tree is the nearest power of 2 which is greater than or equal to the size of the input. A simple illustration is shown in fig. 1. The duplicated points are marked by '*' and '+'. The k-d tree construction provides a representation of each model which is independent of the order in which the points are arranged, as long as the point cloud remains unchanged.\\\\
The provided dataset has models with number of points between 512 and 4096. Hence, for any model, the corresponding order-invariant representation's size can be either 1024,2048 or 4096. For every model to be compatible with a fixed size network, we make sure that each model is represented as a 2048x3 array. For models resulting in 1024x3 arrays, this is achieved by adding a copy of the points with random noise added. For models of size 4096x3, we turn it into two 2048x3 values corresponding to odd and even indexed points.

\begin{figure}[!htb]
\centering
\includegraphics[scale=0.3]{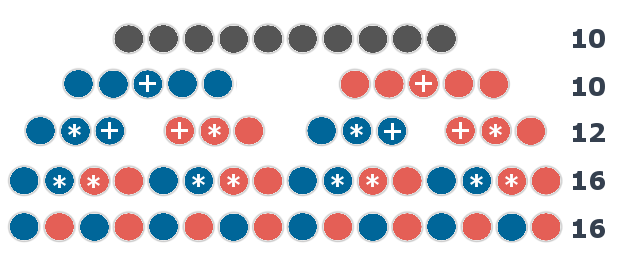}
\caption{k-d tree with point duplication}
\end{figure}

\subsubsection{Segmentation Network}
Segmentation is carried out on the extracted representation. We use fully convolutional networks for this task. One half of the network reduces the spatial size with increase in number of layers. This part is responsible for extraction of high level features. Filters of spatial size 7x3 for the very first layer, followed by 7x1 on the subsequent layers is used. ReLU activations are used in each case. After every two convolutions, a pooling layer is included which reduces the size to a half of the input.\\\\
The shape of input tensor to the network is (2428,3,1) which is the closest value to (2048,3,1) due to restrictions enforced by symmetry in the network. To get the input shape, the actual data of 2048 points is mirrored on the top and bottom to get the extra 380 points. The Conv-Conv-Pool operation is carried out till spatial size of 68x1. The other half of the network, consisting of blocks of De-Convolution, merging (skip connections) and Convolutional layers follows. The network is summarized by the figure below.

\begin{figure}[!htb]
\centering
\includegraphics[scale=0.15]{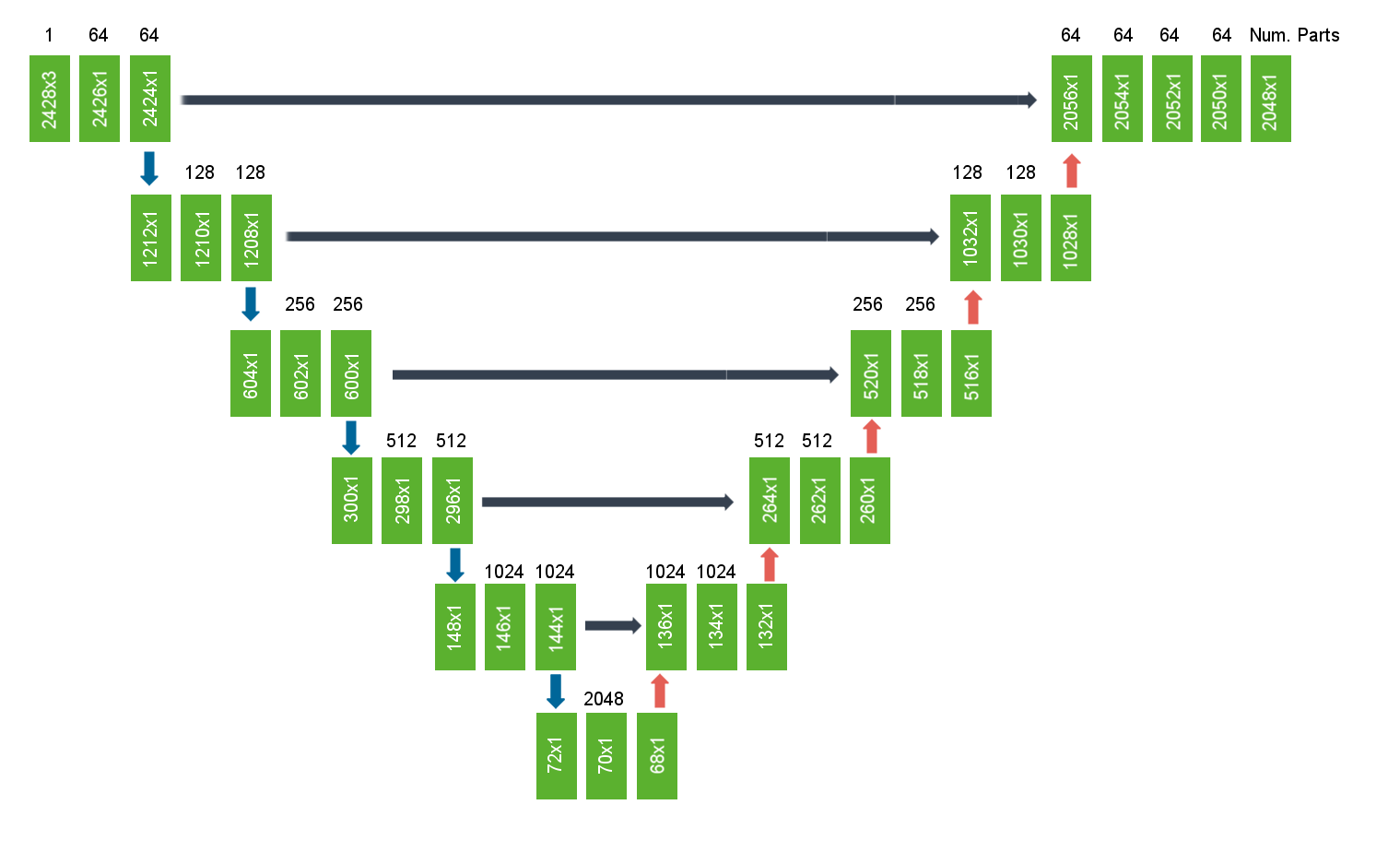}
\caption{Segmentation network architecture}
\end{figure}

Training is done in mini batches of size 32 with cross-entropy loss.

\subsection{DeepPool, by G. Kumar, P. P, S. Srivastava, S. Bhugra, B. Lall}
We utilize the recently proposed PointNet architecture for directly processing the 3D point clouds. The PointNet architecture is augmented with pyramid pooling leveraging skip connections to encode local information. The network is trained end-to-end on the provided training data to achieve the final part segmentation results.
\subsubsection{Method Details}
The overall architecture of the proposed method is shown in Figure \ref{fig:kpsbl_arch}. The initial processing of the point cloud is performed directly on the 3D point cloud (without any voxelization etc.) using the recently proposed PointNet \cite{Qi2017pointnet} architecture. Subsequently, PointNet is augmented with a Pyramid Pooling technique motivated from PSPNet \cite{zhao2017pspnet} which has been recently found to provide excellent results on pixel level labeling in 2D images. 

In the PointNet architecture, we use the features from intermediate layers and augment them with a deep spatial pooling module and train the resulting network end-to-end. Additionally, we perform several modifications to the Spatial Pooling Architecture of PSPNet to suit 3D data. The methodology along with specific contributions is described below:

\begin{figure}[h]
	\begin{center}
		\includegraphics[width=0.5\textwidth, keepaspectratio]{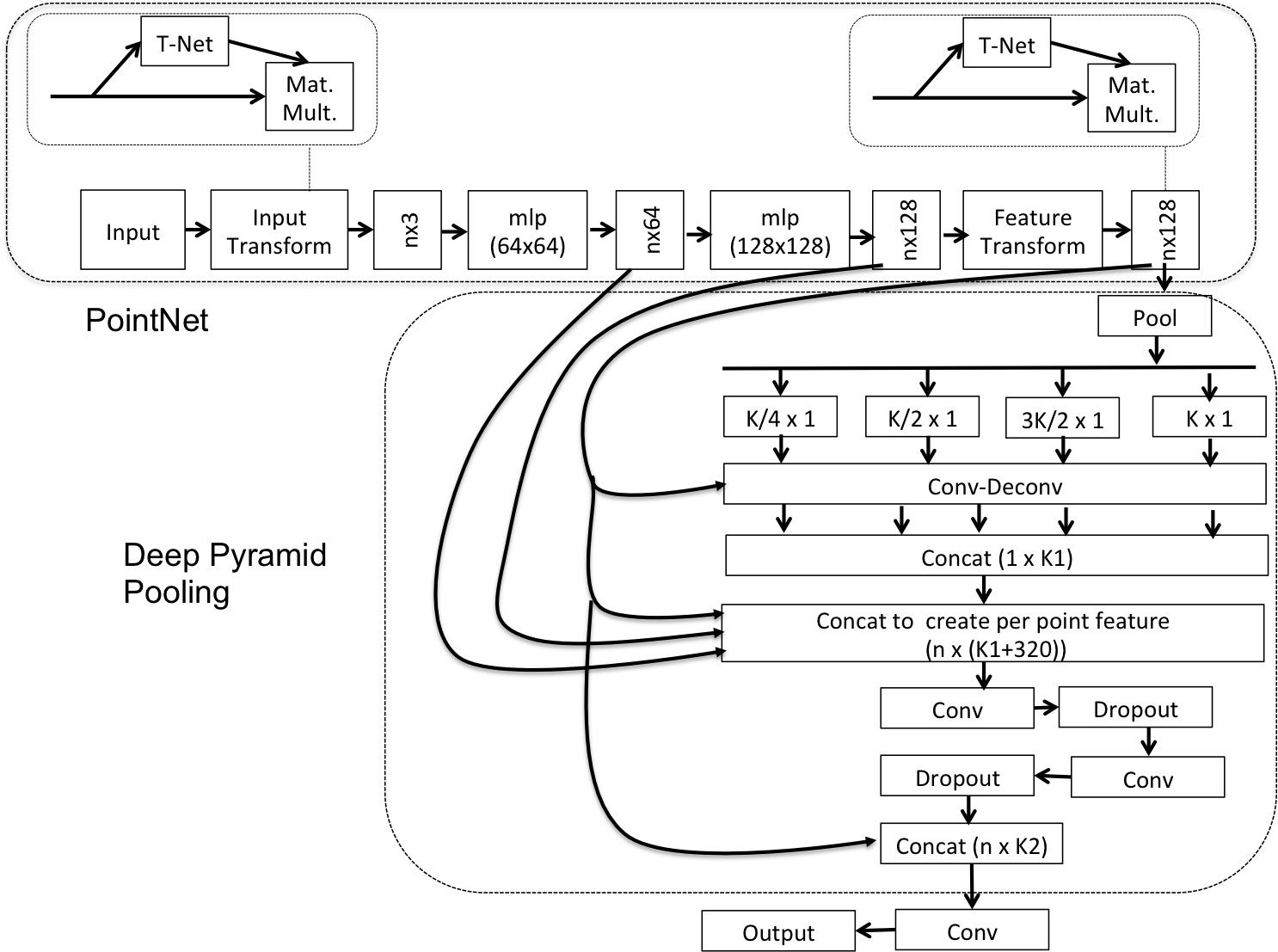}
	\end{center}
	\caption{Overall Architecture}
	\label{fig:kpsbl_arch}
\end{figure}

\begin{itemize}
	\item Like its 2D counterpart, we form four levels of pyramid pooling. However, instead of directly convolving them and upsampling them using bilinear interpolation, we form a network of convolution and deconvolution (Conv-Deconv in Figure \ref{fig:kpsbl_arch}) to get upsampled feature maps. Also, we add another layer of mlp in PointNet architecture prior to pooling the features. The intuition being that in later stage we concat these features to exploit characteristic of local regions to strengthen the feature at various levels of spatial pyramid pooling. 
	\item The output of encoder for each level is concatenated to form a global descriptor ($1$x$K1$). Subsequently, we form point features by concatenating this global feature by forming skip connections from PointNet layers. 
	\item Now instead of performing convolution with the obtained feature map, we stack convolution and dropout layers, whose output is concatenated with the final layer of PointNet. This is finally convolved to obtain the final probabilistic vector of point labels.  
\end{itemize}

\section{Reconstruction Methods}
\subsection{Hierarchical Surface Prediction, by C. H{\"a}ne, S.Tulsiani, J. Malik}
\subsubsection{Method Description}
We developed a method called hierarchical surface prediction (HSP)\cite{hane2017hierarchical}, which is formulated around the observation that when representing geometry as high resolution voxel grids, only a small fraction of the voxels are located around the surfaces. With increasing resolution this fraction becomes larger due to the fact that the surface grows quadratically and the volume cubically. To represent the geometry accurately, fine grained voxels are only necessary around the boundary.

\begin{figure}[!htb]
\centering
\includegraphics[scale=0.37]{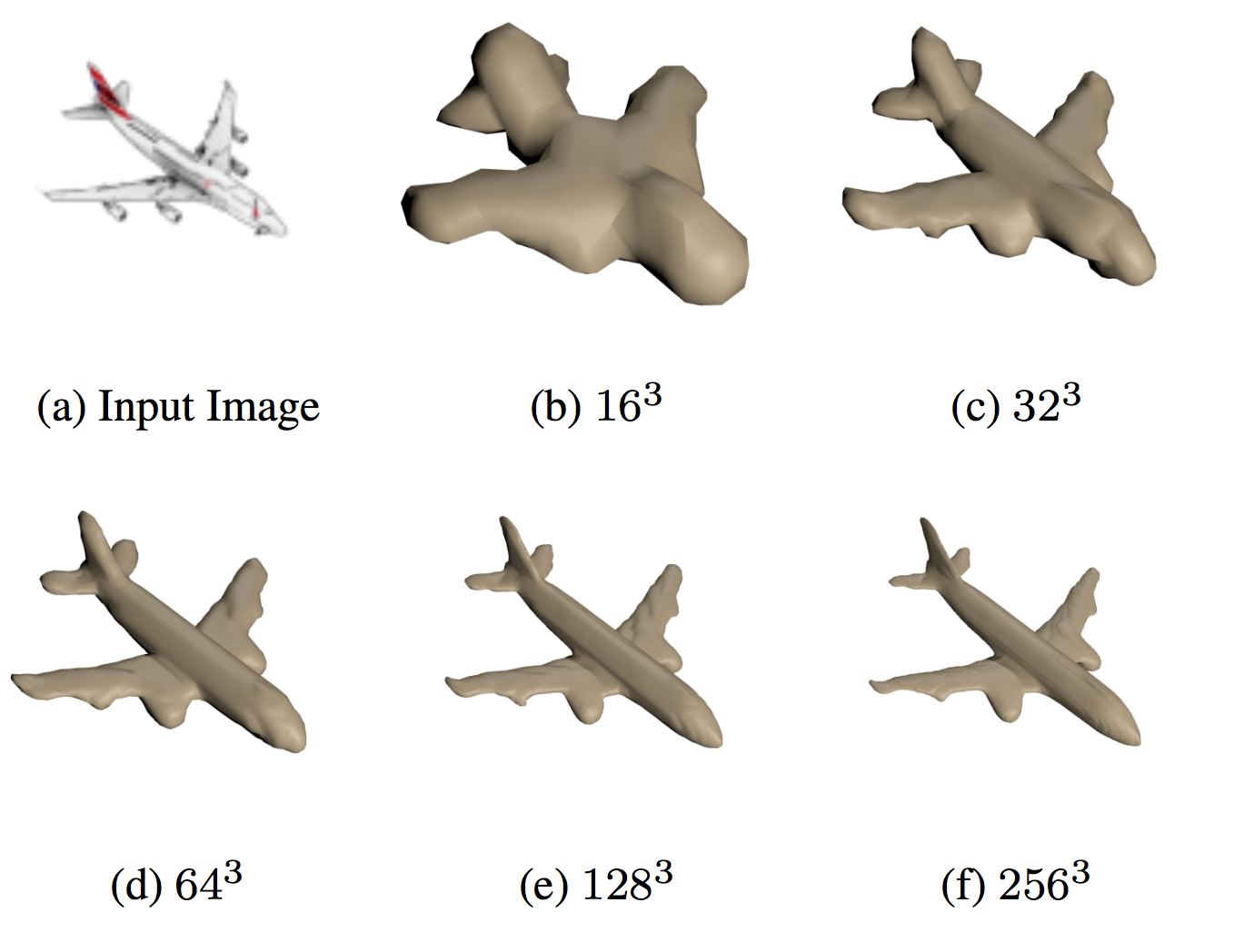}
\caption{Illustration of our method. From the image our method hierarchically generates a surface with increasing resolution. In order to have sufficient information for the prediction of the higher resolution we predict at each level multiple feature channels which serve as input for the next level and at the same time allow us to generate the output of the current level. We start with a resolution of $16^3$ and divide the voxel side length by two on each level reaching a final resolution of $256^3$}
\label{fig:hsp}
\end{figure}

Our method starts by predicting a coarse resolution grid and then hierarchically makes finer resolution predictions. In order to only predict higher resolution voxels where the surface is expected, we change the standard two label prediction to a three label prediction into free space, boundary and occupied space. This allows us to analyze the solution at a specific level and only make finer predictions where the boundary label is present(cf. Figure \ref{fig:hsp}).

In our paper we experimentally show that high resolution predictions are more accurate than low resolution predictions. However, we would like to mention that despite the higher output resolution predicting 3D geometry from a single input image is still an inherently ambiguous task. The focus of our method is not getting the best single view 3D reconstruction results but rather a general framework for high resolution volumetric geometry prediction, and as such our method is not restricted to a specific input modality. For more details about our method see \cite{hane2017hierarchical}.

\subsubsection{Implementation Details}
For this challenge we trained a HSP model with 5 levels $16^3$ , $32^3$ , $64^3$ , $128^3$ and $256^3$. From the predicted voxels we compose a complete solution at $256^3$ voxels by upsampling lower resolution predictions for the locations which have not been predicted at the finest resolution. As input we use the provided RGBA images, i.e. color plus alpha channel. Our network is trained from scratch using Adam. We train for 1 million iterations with a mini-batch size of 4.

\subsection{Densely Connected 3D Auto-encoder by S. Li, J. Liu, S. Jin, J. Yu}
\subsubsection{Method Description}
We present a densely connect auto-encoder to predict 3d geometry from a single image. The auto-encoder consists of three parts: densely connected encoder, full connected layers and densely connected 3D transposed convolutional decoder. All these parts are implemented in a memory efficient way which enables our network to produce high resolution 3d voxel grid. We achieve the results better than 3Dr2n2 and get higher resolution output.

Motivated by a recent study \cite{huang2016densely}, we build our network in a densely connected way. According to the study, dense short connections between standard convolution layers can effectively improve and speed up the learning process for a very deep convolutional neural network. We add 3321 connections in our 82 dense layer 3d auto-encoder and we extend the dense block to 3 dimension which en- ables our network to output 3d voxel grid directly.

\subsubsection{Implementation Details}
We use the densely connected CNN to encode the images into latent vectors. The image encoder DE, which takes a 2d image 128x128 as input and outputs a latent vector z {1024}. The vector contains the information needed to recover a 3d geometry. The dense encoder consists of 5 dense blocks with different number of dense layers {3, 6, 24, 8, 3} and 4 Transition layers which mainly down-sample the feature-maps. As with the encoders, we propose a counterpart decoder which mirrors the encoder. The densely decoder contains the 5 3d dense blocks with{3, 8, 24, 6, 3} dense layer and 4 Up transition layers which mainly up sample the 3d feature-maps. After reaching the output resolution, there will be a final convolution followed by a Sigmoid Layer. The decoder converts the latent vector into the occupancy probability p (i, j, k) of a 3d voxel space.

\begin{figure}[!htb]
\centering
\includegraphics[scale=0.21]{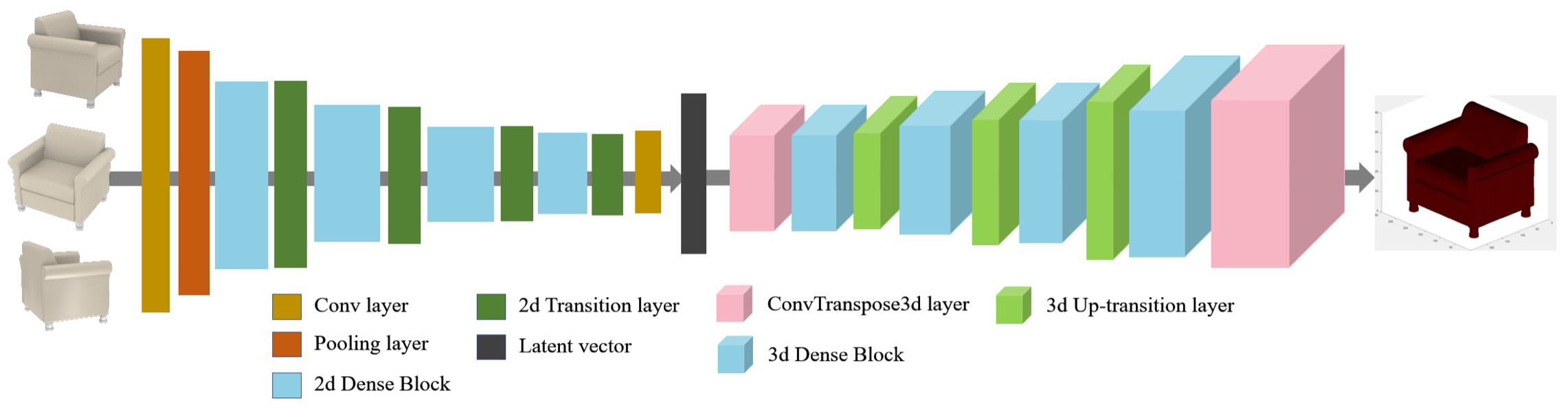}
\caption{Basic architecture for our 3D DCAE}
\label{fig:dense}
\end{figure}

Because of the densely connected pattern, the net will become wider when it grows deeper. This causes the number of parameters to grow quadratically with network depth and with. We implement our dense auto-encoder memory-efficiently based on Shared Memory Allocations, which creates a common place for all layers to store inter- mediate results. Doing so effectively reduce the feature map memory consumption and enables our network output size even up to 128 voxel grid.

The loss of the network is defined as the Binary Cross Entropy between the target and the output. We implement this network using Pytorch. Firstly, we initialize all layers of auto-encoder network. Then, the network was trained for 30 epochs with a batch size of 24. But if output voxel grid is of size 128x128x128, the batch size should be smaller {8} to fit in Nvidia Quadro 4000. We use Relu as activation function and use SGD as optimizer.

\subsection{$\alpha$-Gan, by R.Jones, J.Lafer}
\subsubsection{Method Description}
We extend $\alpha$-GAN\cite{rosca2017variational} to the task of 3D reconstruction from a single image. The network combines a variational autoencoder (VAE) and a generative adversarial network (GAN), where variational inference is accomplished by replacing the KL-divergence with a discriminator and the intractable likelihood with a synthetic likelihood.
\begin{figure}[!htb]
\centering
\includegraphics[scale=0.5]{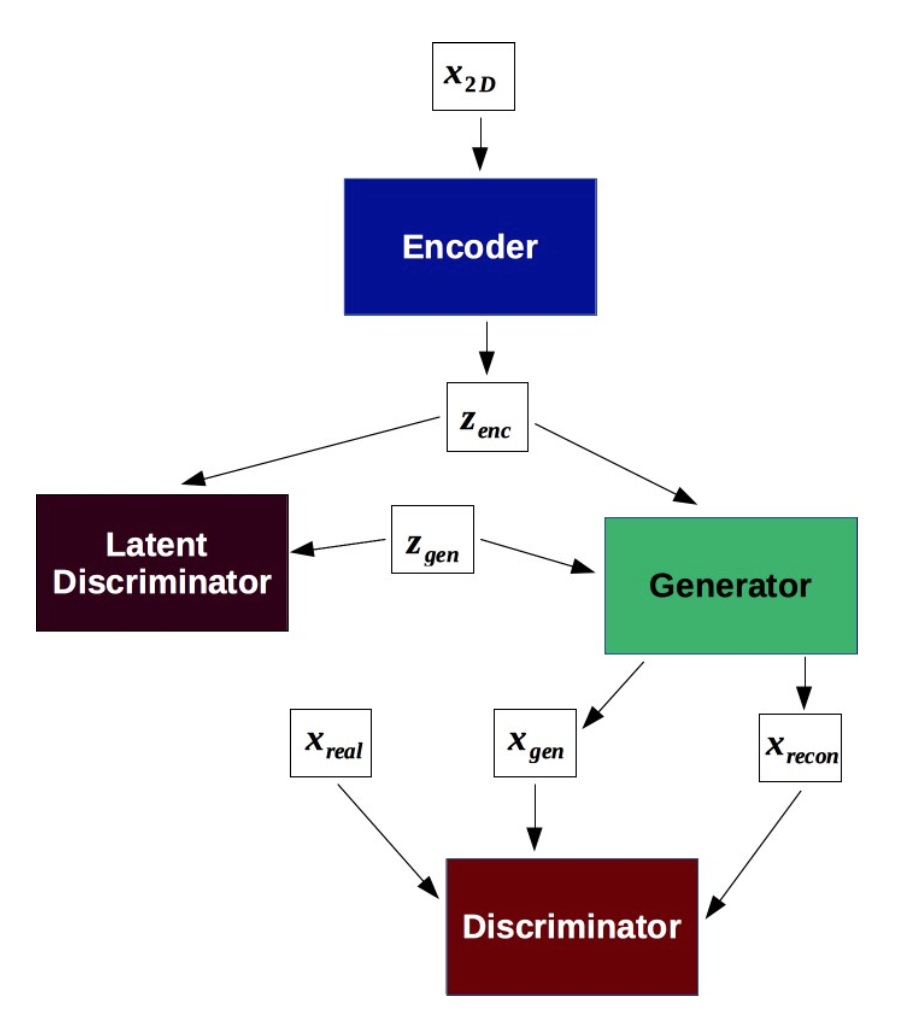}
\caption{Basic architecture for our 3D $\alpha$-GAN}
\label{fig:gan}
\end{figure}
\subsubsection{Model Design}
Pairing a 2D to 3D VAE to a GAN was first explored by Wu et al. \cite{wu2016learning}, and we have enhanced this architecture with a classifier that allows us to treat the variational distribution as implicit. Our model consists of four networks: 1) an encoder (E) that maps 2D images into a latent space; 2) a generator (G) that tries to reconstruct a 3D model from a latent vector; 3) a discriminator
(D)that tries to distinguish real 3D models from reconstructions; 4) a latent discriminator (DL) that tries to distinguish encoded latent vectors from samples of a given distribution. This architecture is demonstrated in Figure \ref{fig:gan}. 
\subsubsection{Training Procedure}
The core training procedure is outlined in Algorithm \ref{alg:a}. The single model was trained over all 55 categories included in the dataset. The voxelizations were downsampled to 64x64x64 from 256x256x256 due to constraints of the hardware.
\begin{algorithm}
  \caption{Pseudocode for $\alpha$-GAN}\label{alg:a}
 \begin{algorithmic}[1]
\State   Let $x_{recon}=\mathcal{G}(z_{enc})$ where $z_{enc}=\mathcal{E}(x_{2D})$ is the latent vector produced from the encoded 2D image, $x_{gen}=\mathcal{G}(z_{gen})$ where $z_{gen}$ is the latent vector sampled from the chosen distribution, and $x_{real}$ is the real training model corresponding to the 2D image.
\For{iter=1:MaxEpoch}
\State Update the encoder and generator by minimizing
$-[x_{real}\log{x_{recon}}+(1-x_{real})\log{(1-x_{recon})}]-\log{\mathcal{D}(x_{recon})}-\log{\mathcal{D}(x_{gen})}-\log{\mathcal{D}_L(z_{enc})}$
\State Update discriminator by minimizing
$-\log{\mathcal{D}(x_{real})}-\log{(1-\mathcal{D}(x_{recon}))}-\log{(1-\mathcal{D}(x_{gen}))}$
\State Update latent discriminator by minimizing
$-\log{\mathcal{D}_{L}(z_{recon})} - \log{(1-\mathcal{D}_L(z_{gen}))}$
\EndFor
\end{algorithmic}
\end{algorithm}

\section{Segmentation Results} \label{seg_results}
\begin{table*}[t!]
\centering
\small
\begin{tabular}{@{}p{0.09\linewidth}|p{0.04\linewidth}|p{0.025\linewidth}p{0.025\linewidth}p{0.025\linewidth}p{0.025\linewidth}p{0.03\linewidth}p{0.025\linewidth}p{0.025\linewidth}p{0.025\linewidth}p{0.03\linewidth}p{0.03\linewidth}p{0.025\linewidth}p{0.03\linewidth}p{0.03\linewidth}p{0.03\linewidth}p{0.03\linewidth}p{0.03\linewidth}}
\hline
method & mean & plane & bag & cap & car & chair & ear-phone & guitar & knife & lamp & laptop & motor-bike & mug & pistol & rocket & skate-board & table \\ \hline
SSCN & \textbf{86.00} & \textbf{84.09} & 82.99 & 83.97 & \textbf{80.82} & \textbf{91.41} & \textbf{78.16} & 91.60 & \textbf{89.10} & \textbf{85.04} & \textbf{95.78} & \textbf{73.71} & \textbf{95.23} & \textbf{84.02} & 58.53 & 76.02 & 82.65 \\
PdNet & 85.49 & 83.31 & 82.42 & 87.04 & 77.92 & 90.85 & 76.31 & 91.29 & 87.25 & 84.00 & 95.44 & 68.71 & 94.00 & 82.90 & \textbf{62.97} & 76.44 & \textbf{83.18} \\
DCPN & 84.32 & 82.75 & \textbf{83.10} & \textbf{87.74} & 76.68 & 89.68 & 73.17 & 91.54 & 86.33 & 80.88 & 95.69 & 67.26 & 95.01 & 80.74 & 62.52 & 74.34 & 82.01 \\
PCNN & 82.29 & 79.19 & 78.53 & 78.90 & 74.56 & 88.82 & 73.20 & 89.11 & 85.98 & 79.51 & 94.53 & 59.14 & 87.95 & 75.94 & 49.09 & 69.51 & 80.32   \\
PtAdLoss & 77.96 & 72.34 & 66.21 & 68.75 & 65.34 & 83.22 & 65.60 & 88.20 & 80.03 & 76.77 & 94.87 & 35.22 & 81.31 & 73.13 & 43.10 & 60.07 & 79.17   \\
KDTNet & 65.80 & 64.34 & 64.88 & 66.69 & 51.42 & 79.26 & 48.96 & 83.93 & 77.80 & 45.56 & 91.65 & 42.50 & 66.14 & 65.15 & 44.36 & 56.75 & 60.00   \\
DeepPool & 42.79 & 19.55 & 34.98 & 27.06 & 5.33 & 40.54 & 8.35 & 27.75 & 49.14 & 18.88 & 14.32 & 9.68 & 34.68 & 52.51 & 22.20 & 16.76 & 78.06   \\ \hline
NN & 77.57 & 78.74 & 71.55 & 69.84 & 69.66 & 83.56 & 48.60 & 88.74 & 79.41 & 74.15 & 93.91 & 61.72 & 88.21 & 78.84 & 50.47 & 66.50 & 72.78 \\
\cite{yi2016syncspeccnn} & 84.74 & 81.55 & 81.74 & 81.94 & 75.16 & 90.24 & 74.88 & \textbf{92.97} & 86.10 & 84.65 & 95.61 & 66.66 & 92.73 & 81.61 & 60.61 & \textbf{82.86} & 82.13 \\ \hline
\end{tabular}
\caption{Summary table of average IoU and per category IoU for all participating teams and methods on the ``original'' test set. Methods are ranked by the average IoU. The last two rows correspond to two baseline approaches.}
\label{tab:seg_results_original}
\end{table*}

\begin{table*}[t!]
\centering
\small
\begin{tabular}{@{}p{0.09\linewidth}|p{0.04\linewidth}|p{0.025\linewidth}p{0.025\linewidth}p{0.025\linewidth}p{0.025\linewidth}p{0.03\linewidth}p{0.025\linewidth}p{0.025\linewidth}p{0.025\linewidth}p{0.03\linewidth}p{0.03\linewidth}p{0.025\linewidth}p{0.03\linewidth}p{0.03\linewidth}p{0.03\linewidth}p{0.03\linewidth}p{0.03\linewidth}}
\hline
method & mean & plane & bag & cap & car & chair & ear-phone & guitar & knife & lamp & laptop & motor-bike & mug & pistol & rocket & skate-board & table \\ \hline
SSCN & \textbf{82.89} & \textbf{75.47} & \textbf{84.04} & 70.58 & \textbf{77.97} & \textbf{88.66} & \textbf{79.54} & 89.45 & \textbf{88.07} & \textbf{83.48} & \textbf{93.26} & \textbf{70.83} & \textbf{95.05} & \textbf{83.28} & 59.98 & 69.53 & 79.87 \\
PdNet & 81.85 & 72.44 & 82.40 & \textbf{78.83} & 72.57 & 87.36 & 77.44 & 88.58 & 86.31 & 82.21 & 91.48 & 64.04 & 93.62 & 81.69 & 62.50 & 68.64 & \textbf{80.69} \\
DCPN & 80.12 & 71.86 & 82.19 & 76.41 & 71.74 & 85.73 & 74.08 & 87.78 & 85.91 & 78.34 & 90.95 & 61.70 & 94.70 & 78.74 & \textbf{62.83} & 67.09 & 79.18 \\
PCNN & 77.14 & 65.46 & 78.81 & 62.92 & 69.77 & 84.39 & 74.66 & 82.45 & 85.65 & 76.71 & 89.49 & 55.31 & 86.87 & 73.59 & 47.85 & 55.54 & 75.86   \\
PtAdLoss & 72.92 & 60.81 & 68.05 & 42.14 & 58.40 & 77.80 & 66.20 & 84.10 & 75.22 & 74.86 & 91.99 & 33.67 & 80.98 & 71.76 & 44.51 & 56.00 & 74.88   \\
KDTNet & 57.71 & 47.06 & 65.60 & 52.87 & 46.53 & 72.70 & 48.51 & 80.06 & 76.70 & 41.20 & 85.27 & 40.17 & 65.93 & 63.09 & 44.17 & 46.70 & 54.26   \\
DeepPool & 43.38 & 33.54 & 37.63 & 22.14 & 4.52 & 37.80 & 8.37 & 26.13 & 45.57 & 19.59 & 15.03 & 9.79 & 34.43 & 51.79 & 22.32 & 13.34 & 75.68   \\ \hline
NN & 70.23 & 63.21 & 69.53 & 40.30 & 62.20 & 75.98 & 48.12 & 84.56 & 77.59 & 71.13 & 88.03 & 54.88 & 87.15 & 75.67 & 48.01 & 54.74 & 66.73 \\
\cite{yi2016syncspeccnn} & 80.85 & 71.55 & 80.61 & 71.31 & 70.65 & 87.03 & 74.62 & \textbf{89.73} & 84.74 & 82.13 & 92.63 & 61.50 & 92.09 & 80.07 & 59.29 & \textbf{75.25} & 78.77 \\ \hline
\end{tabular}
\caption{Summary table of average IoU and per category IoU for all participating teams and methods on the ``deduplicated'' test set. Methods are ranked by the average IoU. The last two rows correspond to two baseline approaches.}
\label{tab:seg_results_deduplicated}
\end{table*}

In this section, we show both per category IoU and the average IoU for each segmentation approach. In addition, we also provide scores for two baseline methods, \cite{yi2016syncspeccnn} and a nearest neighbor based method. The nearest neighbor based method is very straightforward. A most similar shape from the training set is firstly retrieved for each test shape using chamfer distance as the similarity measurement. Since all the shapes are pre-aligned, correspondences can be established via closest point search. Point labels on the test shapes are then predicted through transferring the labels from the training shapes. Segmentation scores on the ``original'' test set and ``deduplicated'' test set are reported in Table~\ref{tab:seg_results_original} and Table~\ref{tab:seg_results_deduplicated} respectively.

Firstly, it is exciting to observe that this year's top submissions outperform previous state-of-the-art by an obvious margin. \textbf{SSCN} has achieved the best performance on both the ``original'' and ``deduplicated'' test sets. All the submissions are based on deep learning approaches and the networks all consume point cloud data directly except for \textbf{SSCN}, which converts input point clouds into voxel representation and train a deep neural network to consume the voxels.

While comparing different categories, it could be seen some categories are more challenging than the others such like motorbike and rocket. These categories usually have smaller number of training data and larger shape variations. All approaches perform consistently worse on these categories. Also we could observe obvious score decline while coming to the ``deduplicated'' test set from the ``original'' test set, indicating the ``deduplicated'' test set is more challenging. After removing the near duplicated models in the test set as well as very similar models to those in the training set, the generalization ability of an approach could be better evaluated. \textbf{DeepPool} restricts the number of predicted parts to be 3 regardless of the category, which severely influences its final evaluation score. Except for it, \textbf{SSCN} achieves the lowest performance drop from ``original'' test set to the ``deduplicated'' test set, showing its better generalizability.

Different 3D representations bring different challenges for deep neural network design. All the participating teams focus on voxel and point cloud representations. Voxel representation is known to be memory expensive in order to achieve high resolution. \textbf{SSCN} leverages sparse convolution to reduce the memory cost while using the voxel representation, achieves the best performance. Point cloud representation loses the regular structure among data points and makes a straightforward extension of grid CNN hard. \textbf{PdNet} uses PCA trees to organize the point clouds and conduct learning on this more structural representation, which leads to the second place among all the submissions. In a similar flavor, \textbf{KDTNet} uses k-d trees to organize the point clouds to make the representation independent of the points order. \textbf{PCNN} introduces the PointConv module to extract point features, which includes ``weighting'' and ``aligning'' operations. It takes the points shape into consideration and is invariant to the points order. The rest submissions are all based upon a previous work, PointNet \cite{Qi2017pointnet}, which is designed to be order invariant to handle learning tasks on a point set. \textbf{DCPN} introduces the idea of dense connections \cite{huang2016densely} into the PointNet architecture. \textbf{PtAdLoss} leverages adversarial loss for shape segmentation based on PointNet and PointNet++\cite{qi2017pointnet++} architecture. \textbf{DeepPool} introduces pyramid pooling on top of PointNet to encode local information, which cannot be well captured by PointNet.

These results indicate that further exploring the properties and impact of different 3D shape representations in deep learning methods would be an interesting direction to go. In addition, though significant improvement has been achieved on shape part segmentation task, there is still large room for improvement.

\section{Reconstruction Results} \label{recon_results}
For shape reconstruction, we use average Intersection over Union (IoU) and average Chamfer Distance (CD)\cite{fan2016point} to evaluate the reconstructed voxels. We provide two baseline methods, a nearest neighbor based method and 3DR2N2\cite{choy20163d}. In the nearest neighbor based method, given a test image, the most similar image in the training set is first retrieved. Then the corresponding training shape is used as the reconstructed shape for the test instance. To get the similarity between two images, we first vectorize the images and then compute the IoU score. For 3DR2N2, we remove the LSTM part in its original network, since for our task 3D voxels are predicted based on a single image. We also downsample the resolution of voxels from $256^{3}$ to $32^{3}$ to run 3DR2N2. The final score of 3DR2N2 are obtained after up-sampling the voxels from $32^{3}$ to $256^{3}$.

All three teams beat the baseline methods by a large margin as is shown in Tabel~\ref{tab:rec_results_deduplicated}. They use different approaches to increase the resolution of the reconstructed voxels. 
\textbf{HSP} achieves highest IoU score on both the "original" and "deduplicated" test sets. Their methods are able to generate voxels at high resolution by hierarchically making finer resolution predictions. \textbf{$\alpha$-Gan} obtains the best CD score on both the "original" and "deduplicated" test sets. CD score cares more about the object geometry and will penalize flying points in a continuous fashion. \textbf{$\alpha$-Gan} achieves the best CD score via leveraging Gan loss. We conjecture this is because Gan loss is more helpful in depicting the geometric correctness than a cross entropy loss. \textbf{DCAE} adopt densely connected neural nets and use Share Memory Allocations to directly predict up to 128 voxel grid.

These results also show that in order to reconstruct voxels at high resolution, the sparsity property of 3D voxels has to be considered. Although large improvement has been made, it still needs more efforts to study how to improve the reconstruction quality of voxels at high resolution.
 
\begin{table}[t!]
\centering
\begin{tabular}{@{}|p{0.22\linewidth}|p{0.115\linewidth}|p{0.13\linewidth}|p{0.14\linewidth}|p{0.14\linewidth}|p{0.15\linewidth}}
\hline
method & IoU(\%) & IoU*(\%) & CD & CD*\\
\hline
HSP &\textbf{38.25}&\textbf{34.71}& 0.003804& 0.004634  \\
$\alpha$-Gan &33.19 &30.89 & \textbf{0.003768} & \textbf{0.004574}\\
DCAE &31.77 & 28.73& 0.005758& 0.007251\\
\hline
NN & 24.24 & 20.50&  0.005514& 0.006677\\ 
3DR2N2($32^3$) & 24.93& 23.04& 0.006627 &0.007604 \\
\hline
\end{tabular}
\caption{Summary table of IoU and CD scores for all participating teams and methods on both the ``original'' and ``deduplicated'' test sets. IoU and CD indicate scores on the ``original'' test set and IoU* and CD* indicate scores on the ``deduplicated'' test set. Methods are ranked by the average IoU score.}
\label{tab:rec_results_deduplicated}
\end{table}

\section{Conclusion}
To facilitate the development of large-scale 3D shape understanding, we establish two benchmarks for 3D shape segmentation and single image based 3D reconstruction respectively on a large-scale 3D shape database. We obtain submissions from ten different teams, whose approaches are all deep learning based. Though the leading teams on two tasks have both achieved state-of-the-art performance, there is still much space for improvement. We have several observations from the challenge: 1. Unlike 2D deep learning, people are still exploring the proper 3D representation for both accuracy and efficiency; 2. What is a good evaluation metric for 3D reconstruction still needs further exploration. We hope researchers could draw inspirations from this challenge and we will release all the competition related data for further research.


{\small
\bibliographystyle{ieee}
\bibliography{egbib}
}

\end{document}